# Triple-Stream Deep Feature Selection with Metaheuristic Optimization and Machine Learning for Multi-Stage Hypertensive Retinopathy Diagnosis

Süleyman Burçin Şuyun 1, Mustafa Yurdakul 2, Şakir Taşdemir 3, Serkan Biliş 4
1 3 Selçuk University, Technology Faculty, Computer Engineering Dept, Konya, Turkey
2 Kırıkkale University, Engineering and Natural Sciences Faculty, Computer Engineering Dept. Kırıkkale, Turkey
3 Selçuk University, Technology Faculty, Computer Engineering Dept, Konya, Turkey
4 Batıgoz Medical Group Hospital, Eye Diseases Dept, İzmir, Turkey

**Abstract;**
Hypertensive retinopathy (HR) is a serious eye disease that can lead to permanent vision loss if not diagnosed early. As traditional diagnostic methods are time-consuming and subjective, there is a growing need for an automated and reliable diagnostic system. The existing studies usually use a single Deep Learning (DL) model and have limited success in distinguishing between different stages of HR. In this study, a three-stage approach is developed to improve the accuracy of HR diagnosis. In the first stage, 14 different well-known Convolutional Neural Network (CNN) models were tested and DenseNet169, MobileNet and ResNet152 were determined as the most successful models. The DenseNet169 model achieved 87.73% accuracy, 87.75% precision, 87.73% recall, 87.67% F1-score and 0.8359 Cohen's Kappa. MobileNet model was the second most successful model with 86.40% accuracy, 86.60% precision, 86.40% recall, 86.31% F1-score and 0.8180 Cohen's Kappa. The ResNet152 model ranked third with 85.87% accuracy, 86.01% precision, 85.87% recall, 85.83% F1-score and 0.8188 Cohen's Kappa. In the second stage, the deep features extracted from the three models were concatenated (feature fusion), and classification was performed using Machine Learning (ML) algorithms (SVM, RF, XGBoost). The SVM (sigmoid kernel) achieved the highest success and the accuracy of the model was 92.00%, precision 91.93%, recall 92.00%, F1-score 91.91% and Cohen's Kappa 0.8930. In the third stage, feature selection was performed with meta-heuristic optimization algorithms (Genetic Algorithm (GA), Artificial Bee Colony(ABC), Particle Swarm Optimization(PSO) and Harris Hawk Optimization(HHO)) to improve classification performance. The HHO algorithm achieved the results of 94.66% accuracy, 94.66% precision, 94.66% recall, 94.64% F1-score and 0.9286 Cohen's Kappa. The proposed method had higher accuracy and generalization values in HR diagnosis than approaches based on a single CNN model and other approaches in the literature.

**Keywords;** Hypertensive retinopathy, Convolutional Neural Network, Feature Fusion, Harris Hawk Optimization, Eye disease

## 1.Introduction

The eyes are one of the most complex and vital organs of the human body, enabling human beings to perceive their surroundings and make sense of visual information by processing it. As the main organ of the sense of sight, the eye functions through the harmonious functioning of sensitive structures such as the cornea, lens, retina and optic nerve[1]. The healthy functioning of these structures directly affects the quality of life of the individual. However, many factors such as genetic factors, aging, infections, chronic diseases such as diabetes, trauma and unhealthy living habits can negatively affect eye health and lead to various diseases[2]. Among eye diseases, serious conditions such as glaucoma, macular degeneration, diabetic retinopathy (DR) and HR are among the leading causes of vision loss worldwide. These diseases can lead to irreversible vision loss if not diagnosed and treated early. Therefore, early diagnosis of eye diseases is critical to protect the quality of life of individuals and to successfully plan treatment processes[3].

Medical imaging techniques have become an integral part of diagnosis and treatment processes in modern medicine. In particular, methods such as Optical Coherence Tomography(OCT) allow diseases to be detected at early stages by imaging sensitive structures of the eye such as the retina, macula and optic nerve with high resolution[4].

However, manual evaluation of these images is both time-consuming and subjective, and is prone to error. In addition, the lack of specialists, especially in rural areas, makes this process even more difficult. For all these reasons, there is a growing need for technologies that automate the diagnosis of eye diseases and make the process both faster and more accessible. Rapid advances in hardware and software technologies have led to the development of numerous Computer Assisted Diagnosis(CAD) systems in this field[5]. In particular, Artificial Intelligence (AI) methods are emerging as effective tools for the diagnosis of eye diseases. There are a number of studies in the literature on AI based automatic diagnosis of eye diseases.

Irshad et al.[6] developed a method that classifies retinal vessels into arteries and veins to calculate the arterio-venous ratio (AVR) for HR detection. In the study, the region of interest (ROI) is extracted by determining the optic disc (OD) center in fundus images and vessels are classified using intensity differences in different color spaces. The average intensity, contrast and roughness features of the vessels in RGB, LAB and HSV color spaces were extracted and classification was performed with support vector machines (SVM). In the tests, the mean density, contrast, and roughness values for arteries were $124.29 \pm 26.09$, $6.40 \pm 1.43$, and $0.0006 \pm 0.0003$, respectively, while for veins these values were $110.61 \pm 27.79$, $9.42 \pm 2.54$, and $0.0013 \pm 0.0007$, respectively. The proposed method was tested on 25 fundus images and an accuracy of 81.3

Abbas et al.[7] developed a system called HYPER-RETINO to classify five stages (normal, mild, moderate, severe, malignant) of HR. The model uses semantic and object-based segmentation techniques to detect HR lesions and classifies them using DenseNet architecture. They tested the model on 1400 fundus images and achieved 92.6% accuracy, 90.5% sensitivity and 91.5% specificity.

Muthukannan [8] proposed a feature extraction-based approach to classify retinal fundus images into age-related macular degeneration (AMD), DR, cataract, glaucoma and normal. They used fundus images from the Ocular Disease Intelligent Recognition (ODIR) dataset. The images were processed using maximum entropy transform to optimize both quality and information content. The feature extraction was performed using a CNN consisting of two convolutional layers and a maximum pooling layer. The hyperparameters of the CNN are optimized with the Flower Pollination Optimization Algorithm (FPOA). After feature extraction, classification is performed using a multi-class SVM. The proposed method achieved 98.30% accuracy, 95.27% recall, 98.28% specificity and 93.3% F1 score.

Kihara et al.[9] developed a ViT-based segmentation model to detect non-exudative macular neovascularization (neMNV) in AMD patients using OCT images. In the study, the model was trained using 125,500 OCT images and the results were compared with human performance. The model The The model achieved 82%, 90%, 79%, and 91% sensitivity, specificity, PPV, and NPV, respectively Compared to human evaluations, the ViT-based model provided higher accuracy and performed particularly strongly in patients with late AMD.

Shoukat et al.[10] performed glaucoma detection on gray channel fundus images using ResNet-50 architecture. Fundus images were processed with grayscale technique and optimized for model training by focusing on the optic disc. The performance of the ResNet-50 model was improved by implementing data augmentation techniques and transfer learning methods. The model was evaluated on the G1020 dataset and achieved high performance results such as 98.48% accuracy, 99.30% sensitivity, 96.52% specificity, 97% AUC and 98% F1-score.

Patel et al. [11] proposed a model combining Flexible Analytic Wavelet Transform (FAWT) and Gaussian-Kuzmin distribution-based Gabor filters (GKDG) to develop an automatic diagnostic tool for the diagnosis of glaucoma. They decompose the green channel components of fundus images using FAWT and extract features from these sub-bands using GKDG filters. Neighborhood Component Analysis (NCA) is applied for dimensionality reduction and the features are classified with LS-SVM algorithm. The proposed model achieved 95.84% accuracy, 97.17% specificity and 94.55% sensitivity in the experiments on the RIM-ONE dataset.

Gu et al.[12] developed a model called Ranking-MFCNet for cataract grading from OCT images. By combining multiscale feature calibration and a ranking-based approach, the model aims to classify six different severity levels of cataracts more accurately. The model uses the eaMFC module, which combines the multiscale attention mechanism with external attention layers to provide a better representation of features to distinguish neighboring severity levels. In the experiments performed on the IOLMaster-700 dataset, the model achieved an accuracy of 88.86%, sensitivity 89.08%, precision 90.15% and F1-Score 89.49%.

Sharma et al.[13] developed a framework for the diagnosis of glaucoma using fundus images. A customized CNN model designed to extract features from the images. In the dimensionality reduction stage, Principal Component Analysis (PCA) and Linear Discriminant Analysis (LDA) techniques were used together. For classification, the Extreme Learning Machine (ELM) method was employed. To enhance the performance of ELM, a Modified Particle Swarm Optimization (MOD-PSO) algorithm was applied to optimize the input weights and biases. The model's performance was evaluated using five-layer cross-validation on the G1020 and ORIGA datasets. The authors obtained success on the G1020 dataset with 97.80% accuracy, 94.92% sensitivity, 98.44% specificity. On the ORIGA dataset, 98.46% accuracy, 97.01% sensitivity, and 98.96% specificity values were obtained.

Kulyabin et al. [14] introduced OCT Dataset for Image Based DL Methods (OCTDL), an open access dataset for DL methods based on OCT imaging techniques. The dataset covers a variety of retinal diseases such as (AMD), diabetic macular edema (DME), epiretinal membrane (ERM), retinal artery occlusion (RAO), retinal vein occlusion (RVO) and vitreomacular interface disease (VID). They also performed experimental studies on OCTDL with ResNet50 and VGG16 models. As a result of these studies, ResNet50 achieved 84.6% accuracy, 89.8% precision, 84.6% sensitivity, 86.6% F1 score and VGG16 achieved 85.9% accuracy, 88.8% precision, 85.9% sensitivity, 86.9% F1 score and 97.7% AUC.

Geetha et al. [15] developed a DL framework for the early detection of glaucoma called DeepGD. For the classification of fundus images, EfficientNetB4 and snapshot ensemble methods were used and Aquila Optimization (AO) algorithm was applied for hyperparameter optimization. The classified images were segmented with V-Net and 99.35% accuracy, 99.04% precision, 99.19% specificity, 98.89% recall, 98.97% f1-score values were obtained.

Liu et al. [16] developed a Swin-TransformerV2-based model for DR diagnosis called STMF-DRNet. The model has a multi-branch structure that allows the extraction of global, local and fine-grained features and aims to improve classification performance by integrating hybrid attention mechanisms and category-based attention mechanisms. In addition, Attention Based Object Localization Module (AOLM) and Attention Based Patch Processing Module (APPM) were implemented to detect lesion regions and reduce noise. The proposed model achieved 77% accuracy, 77% sensitivity, 94.2% specificity, 77% F1-Score, and 87.7% Kappa value on the clinical dataset.

Suman et al.[17] developed a hybrid DL architecture to classify HR severity levels. In the study, HR was classified into four classes (normal, mild, moderate, severe) by creating an expert-labeled HRSG dataset. The model aims to capture both local and global contextual information by combining transfer learning with the pre-trained ResNet-50 and the improved Vision Transformer (ViT) architecture. With the Decoupled Representation and Classifier (DRC) method, class imbalance is removed and the overall diagnostic accuracy of the model is improved. In the tests, the proposed method outperformed existing HR classification methods, achieving 96.88% accuracy, 94.35% sensitivity, 97.66% specificity, 94.42% F1 score and 94.74% accuracy.

**Table 1**.Literature summary of studies on eye disease diagnosis with AI techniques

| Study | Year | Methodology | Imaging Technique | Disease | Results |
|---|---|---|---|---|---|
| [6] | 2014 | SVM-based classification using intensity features | Fundus | HR | Accuracy: 81.3% |
| [7] | 2021 | DenseNet-based semantic segmentation | Fundus | HR | Accuracy: 92.6%, Sensitivity: 90.5%, Specificity: 91.5% |
| [8] | 2022 | CNN-based feature extraction and SVM | Fundus | AMD, DR, Cataract, Glaucoma, Normal | Accuracy: 98.30% Recall: 95.27% Specificity: 98.28% F1-Score: 93.3% |
| [9] | 2022 | ViT-based segmentation model using encoder-decoder architecture | OCT | neMNV | Sensitivity:82% Specificity: 90% PPV: 79% NPV: 91% AUC: 0.91 |
| [10] | 2023 | ResNet-50 architecture with transfer learning and data augmentation | Fundus | Glaucoma | Accuracy: 98.48% Sensitivity: 99.30% Specificity: 96.52% AUC: 97%, F1-Score: 98% |
| [11] | 2024 | FAWT & GKDG based feature extraction and LS-SVM. | Fundus | Glaucoma | Accuracy: 95.84% Specificity: 97.17% Sensitivity: 94.55% |
| [12] | 2024 | A ranking-based multi-scale feature calibration network | OCT | Cataract | Accuracy: 88.86% Sensitivity: 89.08% Precision: 90.15% F1-Score: 89.49% |
| [13] | 2024 | Customized CNN, PCA + LDA for dimensionality reduction, ELM optimized with MOD-PSO | Fundus | Glaucoma | G1020 Dataset: Accuracy: 97.80% Sensitivity: 94.92% Specificity: 98.44% ORIGA Dataset: Accuracy: 98.46% Sensitivity: 97.01% Specificity: 98.96% |
| [14] | 2024 | ResNet50 and VGG16 | OCT | AMD, DME, ERM, RAO, RVO, VID, Normal | ResNet50: Accuracy: 84.6%, Precision: 89.8%, Recall: 84.6%; F1-Score: 86.6%, AUC: 98.8%, VGG16: Accuracy: 85.9%, Precision: 88.8%, |

| | | | | | Recall: 85.9%<br>F1-Score: 86.9%,<br>AUC: 97.7%, |
|---|---|---|---|---|---|
| [15] | 2025 | EfficientNetB4, Snapshot Ensemble, Aquila Optimization and V-Net | Fundus | Glaucoma | Accuracy: 99.35%<br>Precision: 99.04%<br>Specificity: 99.19%<br>Recall: 98.89%<br>F1-Score: 98.97% |
| [16] | 2025 | STMF-DRNet | Fundus | DR | Accuracy:77%<br>Sensitivity: 77%<br>Specificity: 94.2%<br>F1-Score: 77%<br>Kappa: 87.7% |
| [17] | 2025 | Hybrid DL model (ResNet-50 & ViT) | Fundus | HR | Accuracy: 96.88%,<br>Sensitivity: 94.35%,<br>Specificity: 97.66%, F1-Score: 94.42% |

Related studies in the literature are summarized in Table 1. When Table 1 is analyzed, it is seen that many AI-based studies have been carried out in the diagnosis of eye diseases, especially focusing on the detection of diseases such as DR, glaucoma and macular degeneration. However, studies on HR are limited and the literature does not provide a sufficient comprehensive approach in this field. In addition, existing studies have some important limitations. Most of the previous studies, for example, use feature extraction based on a single CNN model. However, a single model may not be sufficient to capture diverse and discriminative features at different stages of HR. Moreover, although some studies integrate machine learning-based classifiers, they do not examine the effectiveness of feature selection methods, which may lead to redundant information content in the extracted features.

Furthermore, although some studies report high accuracy rates, they do not provide a comprehensive analysis of how the model generalizes across different HR severity levels. Whether the model is overfitting, its performance in different patient groups or accuracy rates at specific stages are not examined in detail. These shortcomings indicate the need to develop a more reliable and generalized AI-based system for the diagnosis of hypertensive retinopathy.

The main contributions of this study as follows;
- A comprehensive literature analysis was conducted to examine existing AI-based studies in the diagnosis of eye diseases.
- Fourteen different CNN models commonly used in the literature were trained on a custom HR dataset and the three best-performing models were determined.
- Deep features extracted from the top three CNN models were combined to create a more robust feature set.
- The combined features were classified by ML algorithms (SVM, RF and XGBoost).
- GA, ABC, PSO and HHO methods were used in the feature selection process.
- The classification performance of the model was analyzed using extensive experiments and different evaluation metrics.

The remaining sections of this paper are organized as follows:

Section 2 describes the materials and methods used in this study, including dataset details, DL and ML techniques. Section 3 covers the experimental setup process, including hardware configurations, training procedures and evaluation metrics. Section 4 presents the experimental results, including the performances of CNN models, feature fusion results, and improvements achieved by feature selection. Section 5 discusses the findings and evaluates the strengths and weaknesses of the proposed method. Finally, Section 6 summarizes the main contributions of the work and provides recommendations for future research.

## 2. Material And Methods

In this study, the first step was to prepare the dataset. Then, 14 well-known CNN models were tested and their performances were evaluated. The three most successful models were used as feature extractors and the features obtained from these models were fused(concatenated) and classified with SVM, RF and, XGBoost algorithms. After determining the best-performing ML model, feature selection was performed on the fused features with meta-heuristic algorithms. The details of these methods are explained in the following subsections. Fig. 1 shows the schematic diagram of the proposed three stage approach.

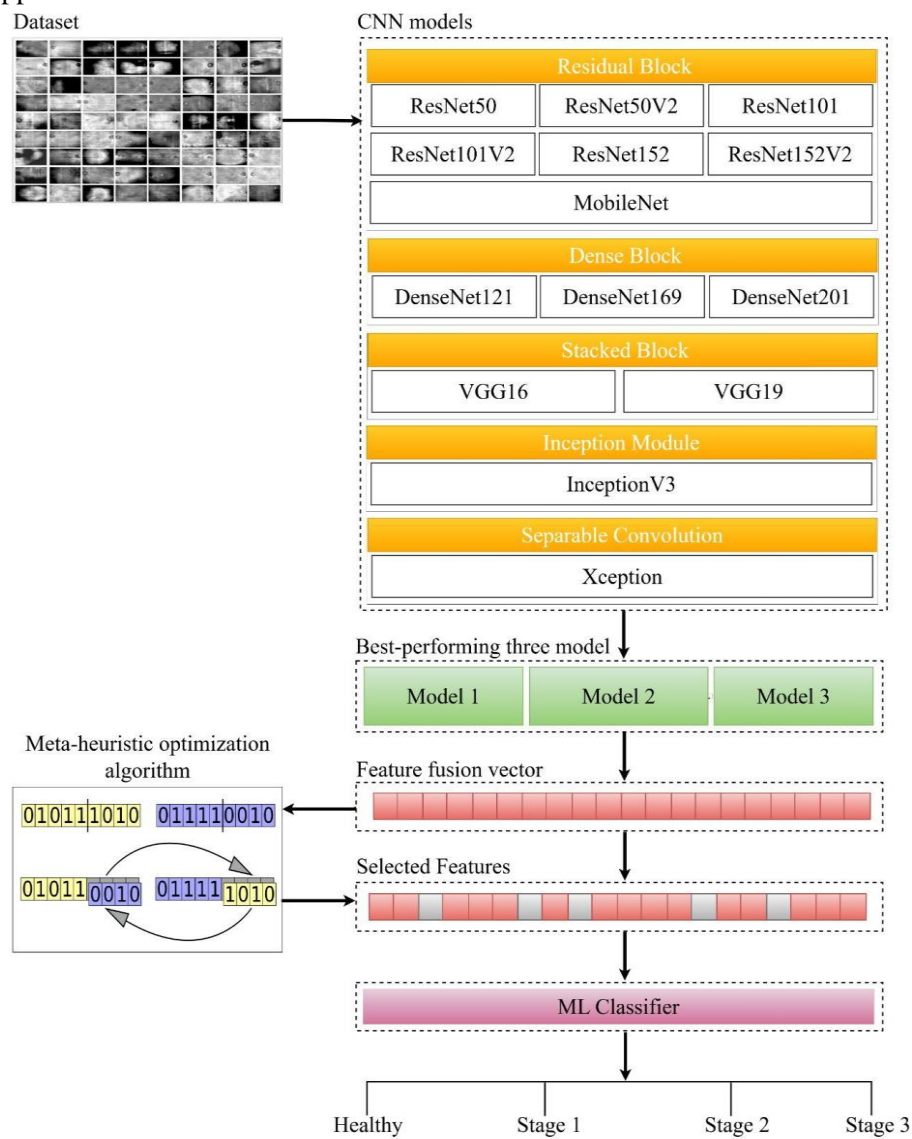

**Fig. 1.** Schematic diagram of proposed three stage approach

## 2.1. Dataset

The HR dataset used in this study consists of specially collected OCT images. The images were obtained from patients at Batıgoz Hospital in İzmir, Turkey, and all patient identifiers were anonymized and protected in accordance with ethical rules. The dataset was created using OCT imaging and manually labeled by an expert ophthalmologist. The labeling process was based on clinical criteria that define the different stages of hypertensive retinopathy. The dataset consists of a total of 1,875 OCT images, and each category is categorized according to a specific stage of the disease.

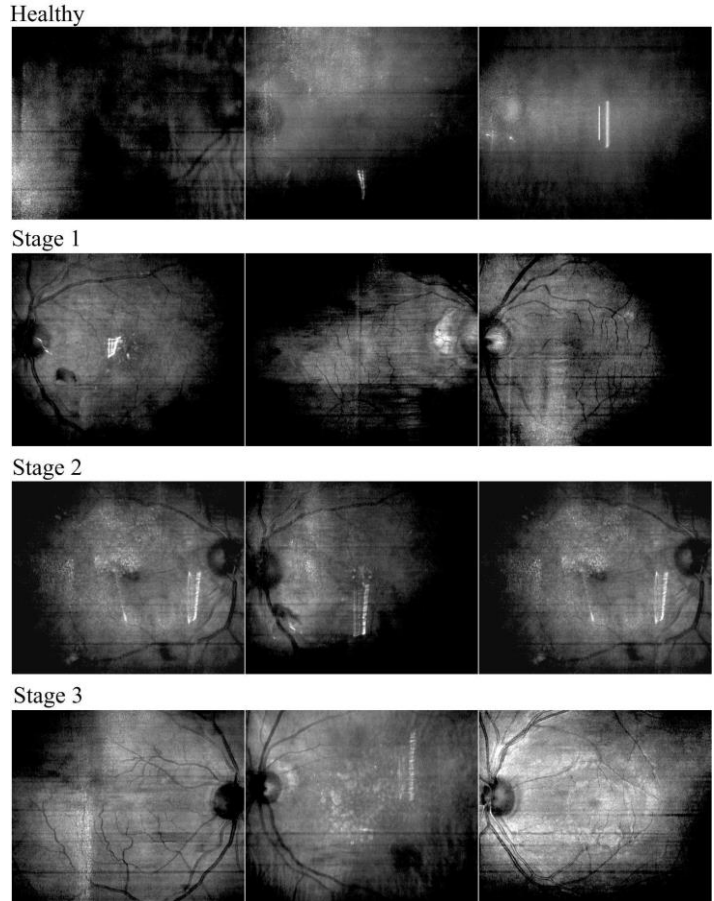

**Fig. 2.** Sample images from HR dataset

Sample images of the dataset are presented in Fig. 2, while the data distribution and the descriptions of the images by category are presented in Table 2. The data distribution is organized to cover different stages of the disease and provides a balanced structure to increase the generalization capacity of the model.

**Table 2.** Image count and identification for each disease category in the HR dataset

| Category | Identification | Image Count |
|---|---|---|
| Normal | No hypertension-related changes or abnormalities in the retina. | 504 |
| Stage 1 | A condition with mild arterial narrowing and thickening of the vessel walls. | 485 |
| Stage 2 | A condition with more pronounced vasoconstriction, arteriovenous crossing and atherosclerosis. | 500 |

| Stage 3 | A condition involving serious vascular disorders with hemorrhages, exudates and cotton wool-like spots in the retina. | 385 |
|---------|---|---|
| Total | | 1875 |

The dataset is divided in a balanced way to train CNN models more efficiently. In this context, 80% of the dataset is used for training purposes, while the remaining 20% is used for testing and validation. In this way, the learning process of the model is optimized while providing sufficient test and validation data for performance evaluation.

### 2.2. CNN

In the field of AI, CNNs are one of the most common and effective methods for analyzing visual data[18]. CNNs are used in complex tasks such as classification[19-21], segmentation[22, 23] and object detection[24-26], and their flexible architecture enables high performance in many different applications. A classic CNN model consists of three basic components: convolution layers, pooling layers and fully connected layers. Convolution layers extract features from the input data through filters, while pooling layers reduce the computational burden by minimizing the size and prevent overlearning. Fully connected layers use these features to perform classification. CNN-based models developed for more complex tasks are optimized for high performance. ResNet[27] solves learning problems in deep networks by using residual learning. DenseNet[28] increases the flow of information with dense connections, connecting each layer to all subsequent layers and preventing information loss. VGG[29] offers balanced performance by stacking small-sized filters on top of each other. Xception[30] reduces computational cost and increases efficiency by using depth separable convolutions. Inception[31] performs robust feature extraction with parallel filters of different sizes. MobileNet[32] is optimized for mobile and embedded devices using depth-separable convolutions, reducing size and computational cost.

### 2.3. Transfer Learning

Transfer Learning is the process by which an AI model adapts previously learned knowledge to a different but similar problem[33]. DL models are often trained on large and diverse datasets, but training a model from scratch for each new problem can be both time consuming and costly. Transfer learning speeds up this process, enabling more efficient results with less data. In this approach, the lower layers of a pre-trained model are usually retained because they have learned general features. The upper layers are retrained specifically for the new problem or completely replaced. In this way, the model can adapt faster to the new data set and achieve high accuracy rates. It is a widely used technique, especially in areas such as image recognition, natural language processing and audio analysis.

### 2.4. ML Algorithms

ML algorithms are models that automatically learn from data and perform certain tasks without human intervention[34]. These algorithms make predictions or decisions by learning patterns from historical data. Basically, it is a structure that allows a system to improve its performance as it gains experience. Some of the most widely used machine learning algorithms are:

*SVM*

SVM[35] is a supervised learning algorithm that tries to find the optimal hyperplane that provides the best separation between classes. It can produce particularly effective results on high-dimensional datasets and performs transformations using kernel functions for non-linearly separable data. The linear kernel is the simplest function used when the data can be linearly separated and aims to separate the hyperplane with maximum margin. The polynomial kernel is used to model nonlinear relationships and increases in complexity as the degree increases. The radial basis function kernel transforms data into higher dimensional spaces, creating complex decision boundaries and is particularly effective with low-dimensional but nonlinearly separable datasets. The sigmoid kernel works similarly to activation functions in neural networks and in some cases can produce similar results to deep learning methods.

*RF*

RF[36] is a robust ensemble learning technique that works as an ensemble of decision trees and is supported by bagging. Each tree is trained on different subsets generated by bootstrap sampling of the training data and using randomly selected feature subsets. The process ensures that each tree captures different patterns; as a result, the predictions of all trees are combined by majority voting or averaging. Thus, the variance of the model is reduced, while the risk of overfitting is significantly reduced. RF can effectively model complex data structures, offering high accuracy and generalization capacity in both classification and regression problems.

*XGBoost*

XGBoost[37] is an optimized version of the gradient boosting algorithm and is known for its high performance, especially on large datasets. A tree-based model, XGBoost builds successive decision trees to reduce errors, with each new tree attempting to correct errors that the previous trees failed to do. Through its regularization techniques and parallel computing capabilities, XGBoost avoids overlearning and runs fast.

In this study, the three ML algorithms mentioned above were used to classify deep features. They were chosen because they are widely used in the literature, offer high success rates and have strong generalization capabilities.

### 2.5. CNN-based feature extraction and fusion

In ML techniques, manual extraction of features is required for image classification. However, this method requires domain knowledge and is difficult to apply especially for complex data patterns. DL methods, on the other hand, eliminate the need for manual feature engineering by learning features automatically. In this study, the best-performing 3 models among 14 different CNN models are identified and the features extracted from these models over the Global Average Pooling (GAP) layer are fused. The feature fusion method aims to create a more comprehensive feature set that can discriminate between classes by combining the features learned by different models. The feature vectors extracted from the three different CNN models are combined using the mathematical formula in Eq. 1

$$F1 = GAP(f_{CNN1}(X)), F2 = GAP(f_{CNN2}(X)), F3 = GAP(f_{CNN3}(X)) \tag{1}$$

$$F_{fusion} = [F1 \parallel F2 \parallel F3] \tag{2}$$

In Eq. 1, $X$ is the input image $f_{CNNi}(.)$ is the feature extraction function of the corresponding CNN model, $GAP(\cdot)$ is the Global Average Pooling process, and $Fi$ is the feature vector extracted from the GAP layer. In Eq. 2, the extracted feature vectors are combined. The operator ∥ denotes the merging of the feature vectors on the horizontal axis. $F_{fusion}$ is classified with the ML algorithms mentioned in Section 2.4. Fig. 3 illustrates the feature fusion process, where deep features extracted from three CNN models are concatenated to create a more comprehensive feature representation for hypertensive retinopathy classification.

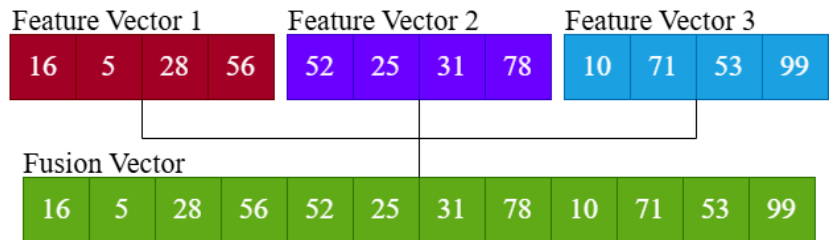

**Fig. 3.** Feature fusion process of deep features extracted from three CNN models

### 2.6. Feature selection with meta-heauristic optimization algorithms

Although feature fusion is a robust method that improves classification performance, some features can cause noise and negatively affect the success of the model[38]. Therefore, determining the optimal feature subset and selecting the variables that best represent the dataset is a critical step. Meta-heuristic optimization algorithms are widely used to improve the accuracy of the model by identifying the most appropriate features in the large search space and reduce the computational cost by eliminating redundant features.

These algorithms try to maximize a fitness function while identifying the best subset of features and at the same time preserving the simplicity of the model, avoiding over-learning. Their flexibility allows them to handle non-linear relationships, interactions between features and problem-specific constraints. Feature selection is usually performed using a binary optimization approach, where each feature is expressed as a binary vector. For a dataset with N features, a binary vector of size N is created, where a value of 0 indicates that the feature is excluded and a value of 1 indicates that the feature is included. The optimization process starts with an initial population or initial solution and is iteratively updated to determine the best feature subset. The binary vector-based feature selection process is shown in Fig. 4.

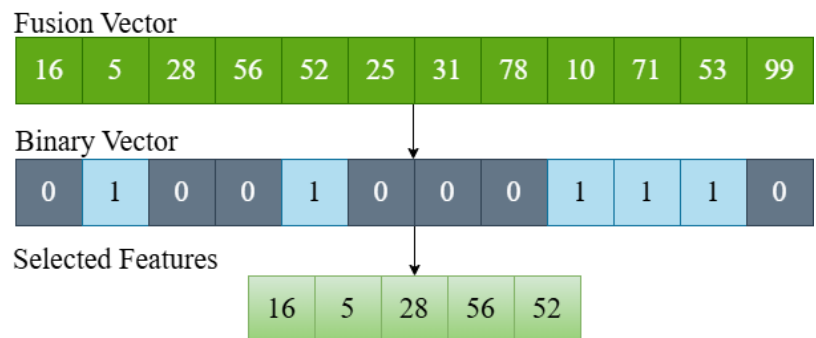

**Fig. 4.** Feature selection process using meta-heuristic optimization algorithms to refine the fused feature set by eliminating redundant features and improving classification accuracy

**Table 3.** The working principle details of the optimization algorithms used in the study

| Optimization approach | Working principle |
|---|---|
| GA | Based on the principles of natural selection and genetic evolution. Solutions are |

|         | improved from generation to generation using genetic operators (selection, crossover, mutation). New solutions are generated by selecting the best individuals and the process continues until the optimum solution is reached. |
|---------|---|
| **ABC** | Mimics the behavior of honeybees searching for food sources. Three types of bees (worker, observer and explorer) try to find the best solutions. While developing good solutions, they also keep discovering new ones. |
| **PSO** | Emulates the collective movements of flocks of birds and schools of fish. Each particle (solution candidate) moves in line with its own best position and the swarm's best position. Velocity updates are based on cognitive (personal best) and social (swarm best) components. |
| **HHO** | Inspired by the hunting strategies of Harris's hawks. It works in a balanced way between exploration (searching for prey) and exploitation (capturing prey), both seeking new solutions and improving existing ones. |

In this study, different approaches based on meta-heuristic optimization were used to select the most suitable features. GA from the evolutionary algorithms group, ABC and PSO from the nature-inspired methods, as well as HHO inspired by predator behavior were evaluated. Table 3 provides details of the optimization algorithms used. These algorithms were selected based on their widespread use in the literature and their successful results. Each of them aims to improve classification performance by using different strategies in the feature selection process, to make the model noise-free and more generalizable.

## 3. Experimental Setup

In all experiments, models are trained in the same hardware and software environment to ensure a fair comparison. The hyperparameters and experimental settings used are standardized to best evaluate the model performance. The details of the experimental setup are detailed in the subsections.

### 3.1. Experiment setting

All experiments in this study were carried out on a system equipped with high-performance hardware and a modern software infrastructure. The details of the experimental environment, including specifications such as the operating system, programming language, framework versions, hardware components, and CUDA version, are comprehensively presented in Table 4.

**Table 4.** Experiment environment specifications

| Configuration | Parameter |
|---|---|
| Operating System | Windows 11, 64 Bit |
| Programming Language | Python version 3.11.4 |
| Fameworks | Tensorflow 2.14.0, Keras version 2.11.4, Matplotlib 3.7.1 |
| GPU | 2 x Nvidia RTX 3090 24GB |
| CPU | Intel(R) Core(TM) i9-10920X CPU @ 3.50GHz |
| RAM | 128 GB |

| CUDA | v12.7 |
|------|-------|

A standardized set of hyperparameters was used to ensure that all DL models were trained under the same conditions. The chosen hyperparameters and their values are presented in Table 5. These parameters are determined using grid-search tecnique.

**Table 5.** CNN model hyperparameters

| Batch size | Epoch | Learning rate | Optimizer | Weights_decay_rate |
|------------|-------|---------------|-----------|--------------------|
| 64 | 100 | 0.0003 | Adam | .0.9 |

### 3.2. Evaluation metrics

The classification performance of DL models can be evaluated by metrics such as accuracy, precision, sensitivity, F1 score and Cohen's kappa coefficient. These metrics are derived from the confusion matrix to analyze the prediction results of the model by class. In the confusion matrix True Positive (TP): iInstances that the model predicts as positive and are actually positive. True Negative (TN): instances that the model predicts as negative and are actually negative. False Positive (FP): instances that the model predicts as positive but are actually negative. False Negative (FN): instances that the model predicts as negative but are actually positive. Accuracy refers to the proportion of the model's total correct predictions. Precision refers to the proportion of positive predicted samples that are actually positive. Sensitivity measures how many true positive samples are correctly predicted. The F1 score provides the balance between precision and sensitivity. Cohen's kappa coefficient assesses how well the model performs compared to random guessing and is calculated based on observed accuracy ($p_o$) and expected accuracy ($p_e$). The mathematical expressions of these performance metrics are shown in Eq. 3-6.

$$Accuracy = (TP + TN) / (TP + TN + FP + FN) \tag{3}$$

$$Precision = TP / (TP + FP) \tag{4}$$

$$Recall = TP / (TP + FN) \tag{4}$$

$$F1 - score = 2(Precision \times Recall) / (Precision + Recall) \tag{5}$$

$$Kappa = (p_0 - p_e) / (1 - p_e) \tag{6}$$

### 4. Results

This section analyzes the experimental results. First, various CNN models were tested on the HR dataset and the three best models were selected based on their classification performance. These models were then used as feature extractors for deep feature fusion. The fused features were classified using different machine learning algorithms to determine the best one. To further improve the performance, meta-heuristic optimization was applied for feature selection and the final model was built using the best machine learning algorithm. Finally, all results are compared and analyzed.

### 4.1. CNN Results

In this section, the performance results of 14 different CNN models tested for HR classification are presented and analyzed. The classification performance of CNN models is provided in Table 6. DenseNet169 was the most successful model with 87.73% accuracy, 87.75% precision, 87.73% recall, 87.67% F1-score and 0.8359 kappa value. The high recall value of the model indicates that patients with HR can be successfully detected. Also, the high precision and F1-score values indicate that the overall accuracy of the model is balanced and the false positive rate is low. The Cohen's kappa value is as high as 0.83, which proves that the model performs much better than a random guessing classifier. MobileNet, the second most successful model, performed quite competitively with 86.40% accuracy, 86.60% precision, 86.40% recall, 86.31% F1-score and 0.8180 kappa value. In particular, the precision value of 86.60% indicates that most of the positive samples predicted by the model are indeed positive.

Table 6. CNN models' classification performance

| Model | Accuracy | Precision | Recall | F1-Score | Kappa |
| --- | --- | --- | --- | --- | --- |
| DenseNet121 | 82.93 | 82.68 | 82.93 | 82.67 | 77.17 |
| **DenseNet169** | **87.73** | **87.75** | **87.73** | **87.67** | **83.59** |
| DenseNet201 | 85.07 | 84.91 | 85.07 | 84.83 | 80.02 |
| InceptionV3 | 82.13 | 81.67 | 82.13 | 81.60 | 76.16 |
| **MobileNet** | **86.40** | **86.60** | **86.40** | **86.31** | **81.80** |
| ResNet101 | 85.07 | 84.88 | 85.07 | 84.92 | 80.01 |
| ResNet101V2 | 85.33 | 85.25 | 85.33 | 85.00 | 80.37 |
| **ResNet152** | **85.87** | **86.01** | **85.87** | **85.83** | **81.88** |
| ResNet152V2 | 83.20 | 83.14 | 83.20 | 82.91 | 77.54 |
| ResNet50 | 84.88 | 84.57 | 84.88 | 84.62 | 79.65 |
| ResNet50V2 | 80.27 | 79.64 | 80.27 | 79.42 | 73.62 |
| VGG16 | 85.87 | 85.38 | 85.07 | 85.06 | 80.83 |
| VGG19 | 85.87 | 85.64 | 85.87 | 85.41 | 81.08 |
| Xception | 84.27 | 84.18 | 84.27 | 84.13 | 78.95 |

The third best performing model, ResNet152, has a very high success rate with 85.87% accuracy, 86.01% precision, 85.87% recall, 85.83% F1-score and 0.8188 kappa value. Especially the precision value of 86.01% shows that the model keeps the false positive rate low. The recall value of 85.87% indicates that most of the HR cases were correctly detected. The Kappa coefficient of 0.81 proves that the model has a statistically significant success and provides significantly better results compared to random prediction.

The models that performed less well include DenseNet121, ResNet101, VGG16, VGG19, Xception and InceptionV3. In particular, VGG16 and VGG19 showed acceptable results with an accuracy of 85.87%, but exhibited a more uneven distribution compared to models with lower precision, recall and F1-score values. The fact that the VGG architecture is older and requires more parameters compared to advanced

CNN architectures limited the generalization ability of the model and caused it not to be as successful as expected on the HR dataset.

Remarkably, the InceptionV3 and Xception models performed lower than expected with 82.13% and 84.27% accuracy rates. Although these models are known for their ability to extract features at different scales, the low recall and F1-score values suggest that the generalization ability of the model is weakened by overfitting in certain classes. Especially Xception, despite its 84.27% accuracy rate, shows an uneven distribution in precision and recall values, indicating that the model is prone to misclassification.

Fig. 5 shows the confusion matrix of the three most successful CNN models. For 0 (Healthy), all models have a high accuracy rate. For 0(Healthy), DenseNet169 predicted 91 correct predictions, while MobileNet and ResNet152 predicted 92 and 90 correct predictions for 0(Healthy), respectively. For Stage 1, the errors are more pronounced, with MobileNet and ResNet152 often misclassifying this class as 0 or 2. DenseNet169 made 73 correct predictions, while MobileNet and ResNet152 made 69 correct predictions and misclassified 23 instances as 0. For Stage 2, DenseNet169 made 89 correct predictions, while MobileNet and ResNet152 correctly predicted 87 instances and misclassified 10-12 instances as 1. The models made the most errors between classes 1 and 2. In the Stage 3 class, all models correctly predicted 76 instances, while only 1-2 instances were incorrectly predicted. Overall, DenseNet169 gave the most consistent results, while MobileNet and ResNet152 made more errors, especially in classes 1 and 2.

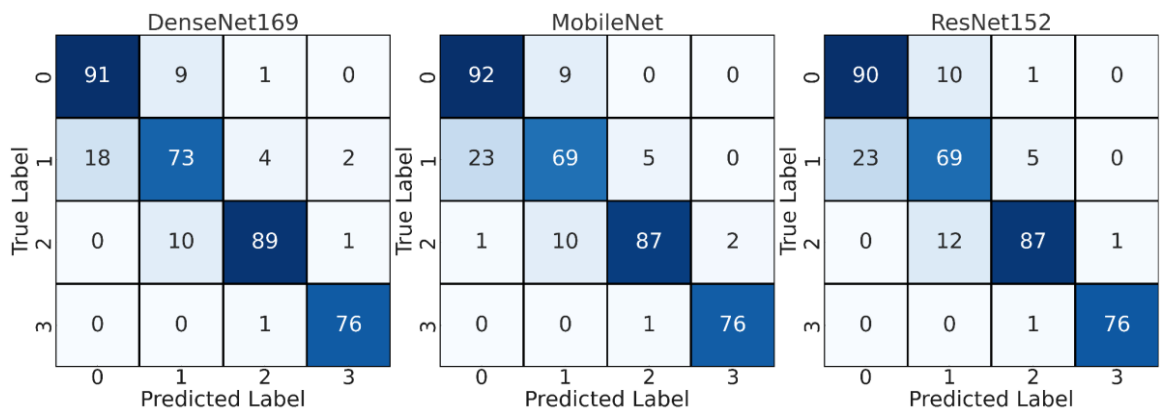

**Fig. 5.** Confusion matrices of three best-performing three CNN models(0: Healthy, 1: Stage 1, 2: Stage 2, 3: Stage 3)

As a result of the overall comparison of CNN models, DenseNet169, MobileNet and ResNet152 were determined as the CNN models with the highest accuracy rate for HR diagnosis. In the next stage, these three models were used as feature extractors to extract deep features and combined with the feature fusion method to further improve the classification performance.

**4.2. Feature Fusion Results**

In this section, the three most successful models, DenseNet169, MobileNet and ResNet152, were used as feature extractors and the features obtained from the three different models were combined by feature fusion. Then, these fused feature vectors were classified by SVM, RF and XGBoost algorithms. The classification performances of the ML algorithms are presented in Table 7.

**Table 7.** Feature fusion results

| Model | Accuracy | Precision | Recall | F1-Score | Kappa |
| --- | --- | --- | --- | --- | --- |

| SVM (linear)  | 88.77 | 88.82 | 88.77 | 88.72 | 0.8497 |
| SVM (polynom) | 91.14 | 91.16 | 91.14 | 91.14 | 0.8818 |
| SVM (rbf)     | 90.11 | 90.16 | 90.11 | 90.05 | 0.8677 |
| SVM (sigmoid) | 92.00 | 91.93 | 92.00 | 91.91 | 0.8930 |
| RF            | 87.97 | 87.98 | 87.97 | 87.91 | 0.839  |
| XGBoost       | 88.60 | 88.52 | 88.59 | 88.62 | 0.8480 |

**Fig. 6.** Confusion matrix of SVM model with sigmoid kernel(0: Healthy, 1: Stage 1, 2: Stage 2, 3: Stage 3)

The experimental results show that the SVM model achieves the highest performance when the sigmoid kernel function is used. SVM (sigmoid) outperformed other machine learning models with 92.00% accuracy, 91.93% precision, 92.00% recall, 91.91% F1-score and 0.8930 kappa value. Notably, SVM's polynomial and RBF kernels also provided high accuracy but were inferior to the sigmoid function. The SVM (polynomial) model achieved 91.14% accuracy, while the SVM (rbf) model achieved 90.11% accuracy.

When other machine learning algorithms were analyzed, the RF model showed a relatively low performance with 87.97% accuracy, 87.98% precision, 87.97% recall and 87.91% F1-score. The XGBoost model performed slightly better than RF with an accuracy of 88.60%. In general, the SVM (sigmoid) model achieved the highest performance in terms of both accuracy and other evaluation metrics and was selected as the best machine learning model.

The results of this stage show that combining features from different CNN models improves classification performance and provides better results compared to traditional single model approaches. Therefore, in the next stage, feature selection was performed using meta-heuristic optimization algorithms and classification was performed with the best machine learning model.

The confusion matrix in Fig. 6 shows the classification performance after feature fusion with the SVM (sigmoid) model. Compared to the individual CNN models, the accuracy increased in all classes. In the 0 (Healthy) class, with 95 correct predictions, the success rate was higher than the previous models. In the 1 (Stage 1) class, 80 correct predictions were made, while misclassifications decreased. Especially in the 2 (Stage 2) class, the best result was achieved with 93 correct predictions and a significant improvement compared to the individual CNN models. In the 3 (Stage 3) class, the highest accuracy was achieved with 77 correct predictions. In general, the SVM (sigmoid) model after feature fusion produced better results than the performance of CNNs alone and made the discrimination between classes more successful.

### 4.3. Feature Selection Results

In this section, meta-heuristic optimization algorithms are used to select the most important features from the large feature set obtained after feature fusion and to improve classification performance. GA, PSO, ABC and HHO algorithms were used for feature selection and the best method was determined. Experimental results show that the HHO algorithm provides the highest classification accuracy. After feature selection using HHO and SVM (sigmoid) model achieved 94.66% accuracy. This result shows that the selection made after feature fusion increases the generalization ability of the model and significantly improves the classification accuracy. When the results obtained with other optimization algorithms are compared, GA achieved 93.23% accuracy, ABC algorithm achieved 93.72% accuracy and PSO achieved 93.23% accuracy. The higher accuracy of HHO indicates that it is the most appropriate feature selection strategy for HR classification. In general, feature selection with meta-heuristic optimization algorithms improved classification success by removing redundant or low-impact features from the model. The HHO algorithm was determined to be the best method as it provided the highest accuracy and consistency and was used in the final model.

**Table 8.** Feature selection performance of different meta-heuristic optimization techniques

| Model | Accuracy | Precision | Recall | F1-Score | Kappa |
| --- | --- | --- | --- | --- | --- |
| GA | 93.23 | 93.27 | 93.23 | 93.24 | 0.909 |
| ABC | 93.72 | 93.73 | 93.72 | 93.72 | 0.916 |
| PSO | 92.96 | 92.98 | 92.96 | 92.97 | 0.9061 |
| HHO | 94.66 | 94.66 | 94.66 | 94.64 | 0.9286 |

**Fig. 7.** Confusion matrix of SVM classification of features selected with HHO

The confusion matrix obtained after feature selection with the HHO algorithm shows a significant improvement in classification performance compared to the SVM (sigmoid) model used in the feature fusion stage. Especially in the 0 (Healthy) class, while 95 correct predictions were made in the feature fusion stage, this number increased to 96 with HHO and the number of incorrect predictions decreased. Similarly, in the 1 (Stage 1) class, while 80 correct predictions were made with feature fusion, this number increased to 86 with HHO and incorrect predictions decreased significantly. One of the areas where the model showed the greatest improvement was in the 2 (Stage 2) class. While 93 correct predictions were made in this class in the feature fusion phase, this value increased to 96 with HHO and the rate of incorrect predictions decreased. Especially in the feature fusion stage, the errors made between classes 1 and 2 were minimized by feature selection. In the Stage 3 class, 77 correct predictions were made with feature fusion and this value did not change with HHO, which shows that the model already classifies the most advanced stage diseases with high accuracy. In general, feature selection with the HHO algorithm provided better discrimination, especially in classes 1 and 2, and increased the overall accuracy of the model by reducing false predictions. Although using all features in the feature fusion stage improves the performance of the model, filtering out unnecessary features allows the model to generalize better and the distinction between classes becomes clearer. These results show that optimization with HHO is the best method for HR classification and the model provides higher accuracy.

### 4.4. Comparison of experiments

In this section, the results obtained at different stages of the proposed method are analyzed comparatively. First, the classification results using only CNN models are analyzed. DenseNet169, MobileNet and ResNet152 were identified as the three most successful models, with DenseNet169 having the highest accuracy rate. However, the classification performance of CNN models alone could not fully resolve the ambiguities between classes.

**Table 9.** Comparison of classification performance across different stages of the proposed method

| Technique | Accuracy | Precision | Recall | F1-Score | Kappa |
|---|---|---|---|---|---|
| DenseNet169 | 87.73 | 87.75 | 87.73 | 87.67 | 0.8359 |
| MobileNet | 86.40 | 86.60 | 86.40 | 86.31 | 0.8180 |
| ResNet152 | 85.87 | 86.01 | 85.87 | 85.83 | 0.8188 |
| Feature Fusion *SVM(Linear)* | 92.00 | 91.93 | 92.00 | 91.91 | 0.8930 |
| Feature Selection *(HHO)* | 94.66 | 94.66 | 94.66 | 94.64 | 0.9286 |

In the feature fusion stage, the features extracted from DenseNet169, MobileNet and ResNet152 were fused and classified with different machine learning algorithms. The results showed that the SVM (sigmoid) model achieved the highest accuracy. At this stage, the accuracy rate increased to 92.00%, which is a significant improvement over methods using only individual CNN models. In the feature selection phase, redundant and low impact features were filtered out using meta-heuristic optimization algorithms. After feature selection with the HHO algorithm, the classification accuracy rate reached 94.66% and the overall performance of the model reached the highest level. Especially in classes 1 and 2, misclassifications decreased and the generalization ability of the model improved. When the overall comparison is made, there is a significant improvement from the results obtained with CNN models alone to the feature fusion stage, and the greatest success is achieved after feature selection. This process shows that it is critical to use both deep learning and optimization techniques together to achieve the best performance in HR classification.

## 5. Discussion

In this study, a robust model for HR diagnosis is developed by combining CNN, machine learning and meta-heuristic optimization methods. The study was carried out in three main stages and the performance of the model was analyzed at each stage.

First, 14 different CNN models were tested and the three highest performing models were identified. DenseNet169, MobileNet and ResNet152 models were used separately for classification and the accuracy rates were calculated as 87.73%, 86.40% and 85.87%, respectively (Table 6). Although the DenseNet169 model had the highest accuracy rate, it was observed that the model could not adequately distinguish between different HR stages and made misclassifications at some stages (Fig. 2). This indicates that approaches based on a single CNN model may be insufficient for HR diagnosis.

In the second stage, the deep features of the three CNN models with the highest accuracy were combined and a feature fusion method was used. The combined feature set was classified with SVM, RF and XGBoost algorithms. The highest accuracy rate of 92.00% was obtained when SVM with sigmoid kernel was used (Table 7). Feature fusion significantly improved the performance of the model compared to individual CNN models. In particular, the misclassifications between stages 1 and 2 of HR decreased and the generalization capacity of the model increased (Fig.6). In the third stage, feature selection was performed using GA, ABC, PSO and HHO to further improve the accuracy of the model. After feature selection, the highest accuracy rate of 94.66% was obtained using the HHO method (Table 8). Through the optimization process, the complexity of the model was reduced, redundant features were filtered out and the overall classification accuracy was improved. In particular, the performance of the model improved by up to 7% in the middle and advanced stages of HR (stages 2 and 3). As shown in Fig. 7, misclassifications were significantly reduced and the model achieved a balanced performance for all classes. As a result of this study, the 87.73% accuracy rate obtained with the single CNN model was increased to 92.00% with deep feature fusion and finally reached 94.66% with meta-heuristic optimization. Thanks to the optimization processes, the accuracy of the model increased by 6.93% and a more reliable system was developed for HR diagnosis.

## 6. Conclusion

In this study, a three-stage model combining deep learning, machine learning and meta-heuristic optimization techniques was developed for hypertensive retinopathy diagnosis. In the first stage, 14 different CNN models were evaluated and the three most successful models (DenseNet169, MobileNet, ResNet152) were selected. However, some classification errors were observed when using CNN models alone. In the second stage, the deep features obtained from these three models were combined and classified with various machine learning algorithms. The best result was obtained by the SVM model with sigmoid kernel with 92.00% accuracy. Feature fusion significantly improved the classification accuracy by increasing the generalization capacity of the model. In the third stage, meta-heuristic optimization algorithms were used to eliminate redundant or low impact features. The HHO method gave the best result compared to other optimization techniques, increasing the accuracy of the model to 94.66%. The improvement made at this stage increased the discrimination between different stages of the disease and reduced misclassifications.

As a result, the developed method provided higher accuracy in diagnosing hypertensive retinopathy compared to approaches based on a single CNN model. Increasing the generalization ability of the model by testing it with different datasets, integrating it into clinical applications and supporting it with explainable artificial intelligence techniques are suggested as important directions for future studies.

**CRediT authorship contribution statement**

**Mustafa Yurdakul, Süleyman Burçin Şuyun:** Conceptualization, Methodology, Review and editing, Software, Validation, Visualization, Writing-original draft, **Şakir Taşdemir:** Conceptualization, Methodology, Supervision. **Serkan Biliş:** Data collection, Data labeling